\def\year{2020}\relax
\documentclass[letterpaper]{article} 
\usepackage{aaai20}  
\usepackage{times}  
\usepackage{helvet} 
\usepackage{courier}  
\usepackage[hyphens]{url}  
\usepackage{graphicx} 
\usepackage{graphicx}  
\usepackage{color}
\usepackage{amsmath}
\usepackage{algorithm, algorithmic,subfig,amsfonts,makecell,array}

\DeclareGraphicsExtensions{.pdf,.jpg,.png}
\graphicspath{{./fig/}}

\def\eg{{\em e.g.}}

\usepackage{threeparttable}
\usepackage{multirow}

\usepackage{amsfonts}

\newcommand{\myPara}[1]{\noindent\textbf{#1}}

\newcommand{\bl}[1]{\textbf{#1}}

\newcommand{\mc}[1]{\mathcal{#1}}

\newcommand{\mr}[1]{\mathrm{#1}}

\newcommand{\bm}[1]{\mbox{\boldmath{$#1$}}}

\title{Look One and More: Distilling Hybrid Order Relational Knowledge for Cross-Resolution Image Recognition}
\author{Shiming Ge\textsuperscript{\rm 1}, Kangkai Zhang\textsuperscript{\rm1,2}, Haolin Liu\textsuperscript{\rm 1,2}, Yingying Hua\textsuperscript{\rm 1,2}, \\ \bf \Large  Shengwei Zhao\textsuperscript{\rm 1,2}, Xin Jin\textsuperscript{\rm 3}, Hao Wen\textsuperscript{\rm 4}\\ 
	\textsuperscript{\rm 1}Institute of Information Engineering, Chinese Academy of Sciences \\
	\textsuperscript{\rm 2}School of Cyber Security, University of Chinese Academy of Sciences\\
	\textsuperscript{\rm 3}Department of Cyber Security, Beijing Electronic Science and Technology Institute~~
	\textsuperscript{\rm 4}CloudWalk Technology Co., Ltd\\ 
	$$\{$$geshiming,zhangkangkai,liuhaolin,huayingying,zhaoshengwei$$\}$$@iie.ac.cn, ~jinxin@besti.edu.cn,~wenhao@cloudwalk.cn 
}
 
\begin{document}

\maketitle

\begin{abstract}
  In spite of great success in many image recognition tasks achieved by recent deep models, directly applying them to recognize low-resolution images may suffer from low accuracy due to the missing of informative details during resolution degradation. However, these images are still recognizable for subjects who are familiar with the corresponding high-resolution ones. Inspired by that, we propose a teacher-student learning approach to facilitate low-resolution image recognition via hybrid order relational knowledge distillation. The approach refers to three streams: the teacher stream is pretrained to recognize high-resolution images in high accuracy, the student stream is learned to identify low-resolution images by mimicking the teacher's behaviors, and the extra assistant stream is introduced as bridge to help knowledge transfer across the teacher to the student. To extract sufficient knowledge for reducing the loss in accuracy, the learning of student is supervised with multiple losses, which preserves the similarities in various order relational structures. In this way, the capability of recovering missing details of familiar low-resolution images can be effectively enhanced, leading to a better knowledge transfer. Extensive experiments on metric learning, low-resolution image classification and low-resolution face recognition tasks show the effectiveness of our approach, while taking reduced models.
\end{abstract}

\section{1~~Introduction}
\begin{figure}[t]
\centering
\includegraphics[width=1.0\linewidth]{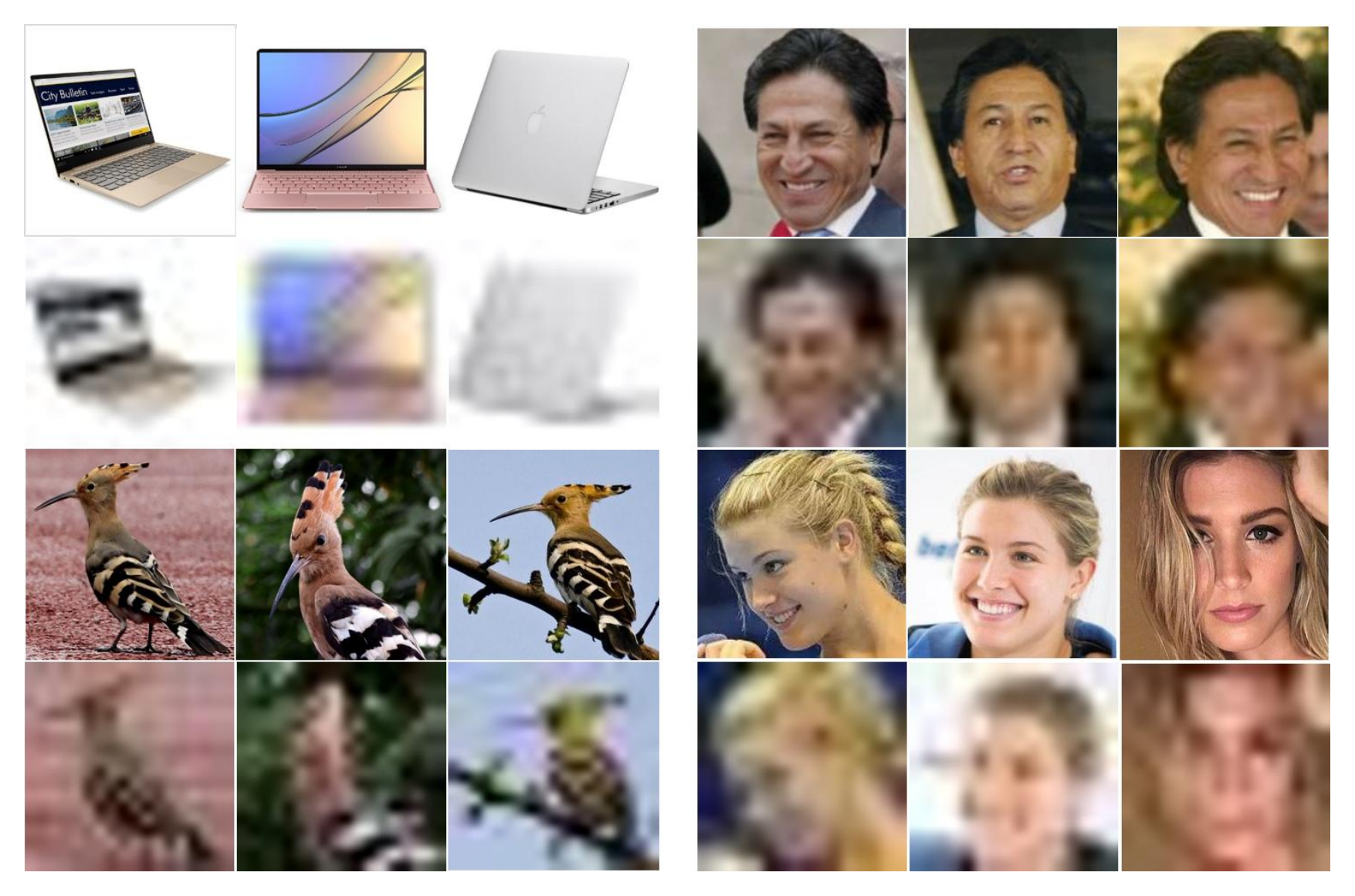}
\caption{Our motivation. When subjects are more familiar with high-resolution images, they can recognize the corresponding low-resolution images well. Thus, by mimicking this capability in recognizing an low-resolution image, we should let the model see more corresponding images.}
\label{fig:motivation}
\end{figure}

With the rapid development of deep learning, recent deep models have proven success in many image recognition tasks. For example, the ResNet model~\cite{he2016cvpr} has achieved a high top-5 accuracy of 96.43\% on ImageNet~\cite{Deng2009ImageNet} object classification task, while VGGFace2~\cite{cao2018vggface2} model gived a 99.63\% accuracy in face verification task on LFW benchmark~\cite{Huang2007report} and ArcFace~\cite{deng2019arcface} reached 98.35\% rank-1 accuracy in face identification task on the challenging MegaFace benchmark~\cite{Kemelmacher2016The}. These successes may arise from that deep models with massive parameters provide an effective way to extract rich knowledge from large-scale data. However, a sharp drop in accuracy may happen when directly deploying these models on recognizing low-resolution images that are difficult to annotate but widely encountered in real-world applications, such as surveillance faces in the wild~\cite{li2019on} and thumbnail images in the Internet. To meet real-world requirements, it is necessary to explore an economic yet feasible solution that can address a key challenge: how to convert an existing complex image recognition model into an efficient one that still works effectively on low-resolution images?

As shown in Fig.~\ref{fig:motivation}, low-resolution images can be recognizable for subjects that are familiar with the corresponding high-resolution images. Intuitively, it is helpful to improve the recognition capacity of a subject by showing more similar or different images as well as more information about their relationships. Thus, the knowledge from high-resolution images can help the extraction of discriminative features for effective recognition. Inspired by that, many models have been proposed to recognize low-resolution images, and can be grouped into two categories: reconstruction-based and representation-based models.

Reconstruction-based models aim at reconstructing the high-resolution images before recognition~\cite{Luan2017Disentangled,2019arXiv190510777K}. These models generally exhibit impressive results in recognizing the reconstructed high-resolution images, but the super-resolution operation often takes additional computational cost. Unlike reconstruction-based models, representation-based models try to extract discriminative features from low-resolution images directly by using various external contexts. For example, ~\cite{biswas2012multidimensional} proposed mapping low-resolution face images into Euclidean space, and then approximating high-resolution face images through distance dimension. In~\cite{ge2019tip},  the authors proposed graph-based optimization algorithm that can extract the most discriminative facial features from existing face models, which can supervise the training process of low-resolution face models. Generally speaking, the most important process in these models is transferring the knowledge from high-resolution images to low-resolution ones. However, it needs to carefully address a key issue in this process: how to represent the knowledge effectively for distilling more information cues and transfer it from high-resolution domain to low-resolution one.

\begin{figure*}[t]
\centering
\includegraphics[width=1.0\linewidth]{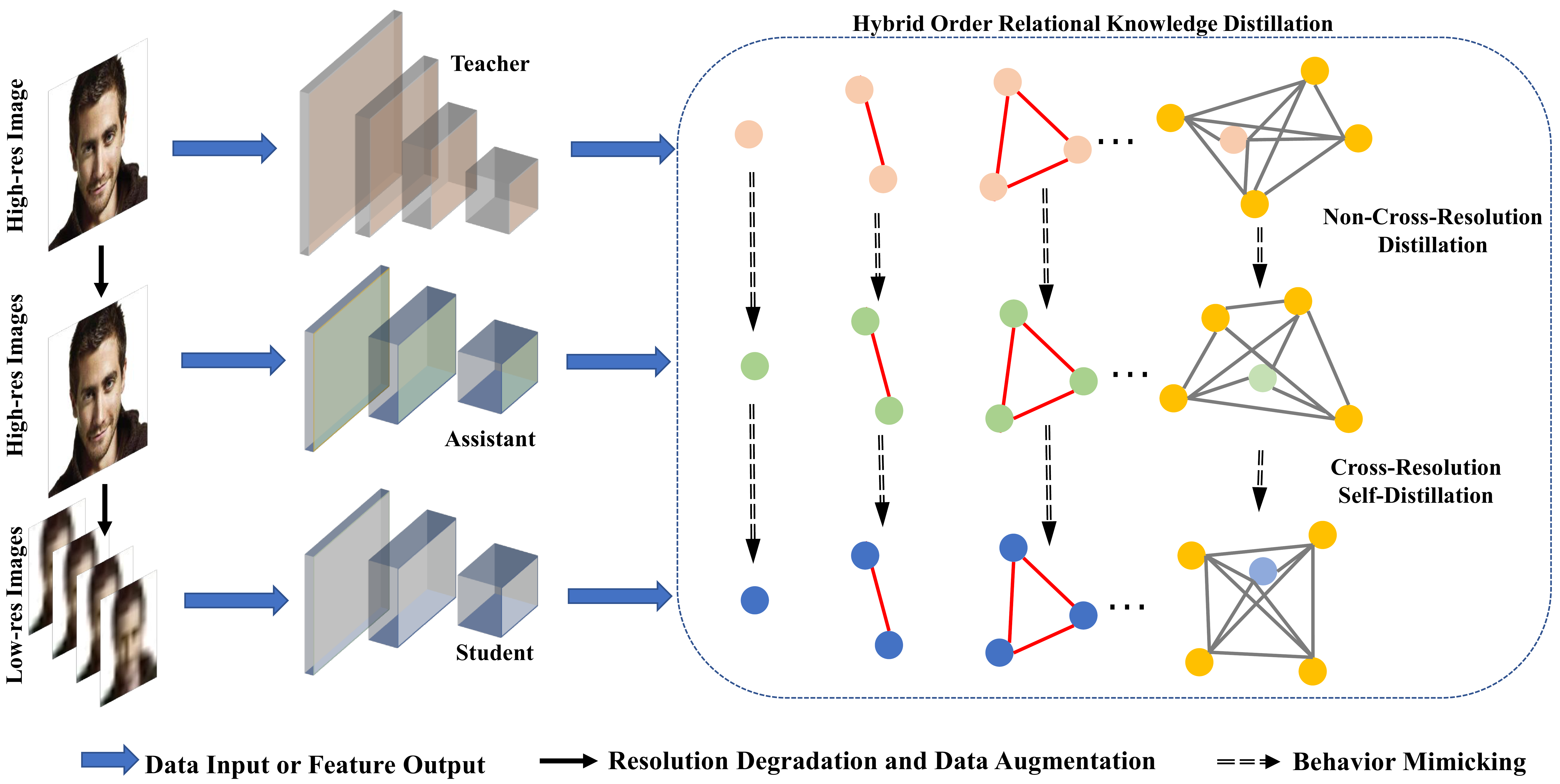} 
\caption{The framework of our approach. By focusing on the structural relational knowledge in various orders, our approach step-wisely transfers knowledge from a complex teacher to a light teacher with an assistant by cross-structure distillation and and cross-resolution distillation.}
\label{fig:framework}
\end{figure*}

Inspired by this fact, we propose a hybrid order relational knowledge distillation approach for low-resolution image recognition. As shown in Fig.~\ref{fig:framework}, the approach consists of a teacher stream, a student stream and an assistant stream. The teacher stream is initialized with pretrained high-resolution image recognition models. Then, the structural knowledge containing one and more order relational information is extracted from the teacher and then transfer to supervise the training of student. In this manner, the student is constructed by showing more information and thus can improve the capacity in recognizing low-resolution images. The assistant with high-resolution image as input helps the student to transfer knowledge when needed. Experimental results on several tasks show that the proposed approach performs impressively in recognizing low-resolution images, with lower memory footprint and faster speed.

Our main contributions are three folds: 1) we propose a hybrid order relational knowledge distillation approach that is able to distill richer knowledge from pre-trained high-resolution models to facilitate low-resolution image recognition; 2) we propose a relation module to extract multiple order relational knowledge; 3) we conduct extensive experiments to show the impressive performance of our approach in metric learning, low-resolution image classification and low-resolution face recognition tasks.

\section{2~~Related Works}

\subsection{2.1~~Low-Resolution Image Recognition}
\myPara{Face Recognition.}
   Low-resolution face recognition has became one of the most difficult problem in the field of face recognition. Low-resolution face image lacks plenty of face details, which results in unsatisfactory accuracy with normal face recognition models. Construction-based and projection-based approaches are proposed to exploit the knowledge of high-resolution images. The construction-based approaches explicitly reconstruct high-resolution facial details from low-resolution images, while the projection-based approaches proposed to hallucinate or super-resolve the high-resolution faces before recognition by explicitly details. Although, these approaches generally achieve good recognition accuracy, they suffer from intensive computation due to the extra face reconstruction process. Another strategy is mapping the features learned from high-resolution images into a domain through transfer learning. The feature extracted by teachers with deeper and more complex networks, which are used to supervise training the student. In order to compensate for the lack of low-resolution image information, high-resolution and low-resolution images are usually input in pair-wise manner. The resulted simplified model can run faster while maintaining good performance and costing reduced memory.

\myPara{Image Classification.}
Compared to low-resolution face recognition, low-resolution image recognition remains less relatively explored. Previous studies show that there have two main ideas to solve the problem. One idea is mapping the source space of low-resolution image to special space and making the largest similarity between high-resolution and low-resolution images. The representative method was proposed by~\cite{wei2017joint}, which found a robust linear function to map different vector space getting the sparse structure of each class. Another idea requires super resolution process, which usually be used to get more detail information about the low-resolution images. For example, ~\cite{wang2016studying} proposed incorporating the super resolution domain adaptation and robust regression step by step. Several works~\cite{jaffe2017super,noor2019gradient} introduced super-resolution for fine-grained low-resolution image classification or other specific low-resolution image classification task. There exist other methods to address low-resolution image classification problem. For example, \cite{singh2019dual} proposed applying dual directed capsule network to low-resolution image classification.

\subsection{2.2~~Knowledge Distillation and Transfer}

With the improvement of model performance, the number of model parameters is also increasing. It often needs to occupy huge memory resources and consume lots of time, leading to the issue of model compression. Knowledge distillation~\cite{2015arXiv150302531H,romero2015iclr,liu2019knowledge,park2019relational,Xue2019CVPR} provides a feasible way to address that issue. Simply speaking, its idea is using a large parameter model (teacher) to supervise the learning of a small parameter model (student). In practice, the student model  learns the output behaviors of the teacher model that has been pre-trained on some target dataset. Although the student model can't reach the accuracy of the teacher model, it still is very powerful and yet efficient. There are two key factors of knowledge distillation: what knowledge has been learned and how to effectively transfer knowledge from teacher model to student model.

Traditional knowledge distillation tends to the direct transmission of instance information, and the teacher and student often input in pairs, losing the structural information about original data. Some recent approaches~\cite{yu2019learning,liu2019knowledge} attempt to pay more attention to the structural relationship of the output rather than the output itself, so when training student model, they try to mimic the same relationship structures as teachers. These approaches can retain higher-order attribute information in transfer learning. Intuitively, more effective information can be obtained by focusing on higher-order relational knowledge, which inspires the original intention of our new distillation approach.

\section{3~~Proposed Approach}
\subsection{3.1~~Framework}
Usually, the higher the order for information we focus on, the more information we can get. Inspired by this and~\cite{DBLP:journals/corr/abs-1902-03393}, we focus on different orders for information attention and propose hybrid order relation knowledge distillation (HORKD) for facilitating low-resolution image recognition (see Fig.~\ref{fig:framework}) via two-stage knowledge transfer. The first cross-structure distillation stage transfers various order relational knowledge from a high-resolution cumbersome model (teacher) $\phi_t(\bm{x};\bm{\mr{w}}_t)$ to a high-resolution compact model (assistant) $\phi_a({\bm{x}};\bm{\mr{w}}_a)$ that mimics the behaviors of the teacher. Then, we transfer various order relational knowledge from $\phi_a(\bm{x};\bm{\mr{w}}_a)$ to a low-resolution compact model (student) $\phi_s(\hat{\bm{x}};\bm{\mr{w}}_s)$ that mimics the behaviors of the the assistant with cross-resolution distillation. Here, $\bm{x}$ and $\hat{\bm{x}}$ are a high-resolution and low-resolution image, respectively. $\bm{\mr{w}}_t$ , $\bm{\mr{w}}_a$ and $\bm{\mr{w}}_s$ are the model parameters.

Traditional knowledge distillation approaches pay more attention to the point-wise relationship between instances in representation space, where the transfer of knowledge may be inadequate. In contrast, our approach aims to achieve better knowledge transfer by considering higher order relations. By redefining the loss function, the student can learn the structural knowledge extracted by the teacher well, and it effectively compensates for the lack of necessary information brought by resolution degradation, thus improving its recognition performance. Toward this end, denoting $\chi^{n}$ and $\hat{\chi}^{n}$ as a set of $n$-order tuple of distinct high-resolution and low-resolution instances respectively, $\bm{f}_i=\phi_t(\bm{x}_i;\bm{\mr{w}}_t)$ as the teacher knowledge distilled from a high-resolution image $\bm{x}_i$ and $\bm{g}_i=\phi_s(\hat{\bm{x}}_i;\bm{\mr{w}}_s)$ as the student knowledge from the responding low-resolution image $\hat{\bm{x}}_i$, the distillation process for $n$-order can be formulated as
\begin{equation}\label{eq:rkd}
\mc{L}_{n}=\sum_{\substack{(\bm{x}_1,...\bm{x}_n)\in{{\chi}^{n}}\\
 (\hat{\bm{x}}_1,...\hat{\bm{x}}_n)\in{{\hat{\chi}}^{n}}} }
\ell (\psi(\bm{f}_1,...\bm{f}_n),\psi({\bm{g}}_1,...{\bm{g}}_n)),
\end{equation}
where $\psi$ is a relational potential function that measures a relational energy of the given $n$-tuple, and $\ell$ is a loss that penalizes difference between teacher and the student.

\subsection{3.2~~Hybrid Order Relational Knowledge}
It is obvious that the relational potential function $\psi$ plays a key role in extracting relational knowledge, which affects the effectiveness and efficiency of knowledge distillation process. Generally, a higher-order potential function may be more powerful in capturing higher-level structure information when costing more computations. Suppose an output representation space (\eg, a mini-batch) has $m$ examples, then the size of a $n$-tuple space is its combination $C_{m}^{n}$.

Our distillation process tries to match the potential energy information between teacher stream and student stream. Therefore, small batch normalization is very useful, especially when the difference between the two streams is significant.
In order to compensating the information loss in resolution degradation, we expect to transfer various order relational knowledge. Therefore, it needs to exploit an effective solution to address this in an efficient way. In this work, we first employ the traditional low-order relational knowledge: 1-order, 2-order and 3-order. Furthermore, we propose an effective approach to exploit higher-order relational knowledge.

\myPara{1-order relational knowledge.} It is also known as point-wise distillation loss, when $n=1$ and the relation is unary. It is popular with previous works, which uses the class probabilities produced from the teacher as soft targets for training the student or transferring the intermediate feature maps. In this work, we transfer features and the loss function is given as follows,
\begin{equation}\label{eq:rkd-1}
\mc{L}_{1}=\sum_{\bm{x}_i\in{{\chi}^{1}}, \hat{\bm{x}}_i\in{{\hat{\chi}}^{1}}} \ell_1(\bm{f}_i,{\bm{g}}_i),
\end{equation}
where $\ell_1$ is a function which measures the $\mc{L}_{1}$ distance between the feature instances of teacher and student.

\myPara{2-order relational knowledge.} It is also known as pair-wise distillation loss. Recent works have used it in various tasks such as image classification, image retrieval and semantic segmentation. Its objective is transferring pair-wise relations, specially pair-wise similarities in our approach, among instances. We adopt the square difference to formulate the 2-order relational knowledge distillation loss,
\begin{equation}\label{eq:rkd-2}
\mc{L}_{2}=\sum_{\substack{(\bm{x}_i,\bm{x}_j)\in{\chi^2} \\ (\hat{\bm{x}}_i,\hat{\bm{x}}_j)\in{\hat{\chi}^2}}} \ell_2(\psi_d(\bm{f}_i,\bm{f}_j),\psi_d({\bm{g}}_i,{\bm{g}}_j)),
\end{equation}
where $\psi_d$ is a normalized loss function that measures the L2 distance between the features from two instances in a mini-batch space and having $\psi_d(\bm{f}_i,\bm{f}_j)=1/\mu\|\bm{f}_i-\bm{f}_j\|_2$, where $\mu=1/|\chi^2|\sum_{(\bm{x}_i,\bm{x}_j)\in{\chi^2}}{\|\bm{f}_i-\bm{f}_j\|_2}$.

\myPara{3-order relational knowledge.} It measures the relation among the examples in a triplet. Toward this end, Park $et$ $al.$ propose an angle-wise distillation loss that is formed by three examples in the output feature space:
\begin{equation}\label{eq:rkd-3}
\begin{split}
\mc{L}_{3}=\sum_{\substack{(\bm{x}_i,\bm{x}_j,\bm{x}_k)\in{\chi^3} \\ (\hat{\bm{x}}_i,\hat{\bm{x}}_j,\hat{\bm{x}}_k)\in{\hat{\chi}^3}}}
\ell_3(\psi_a(\bm{f}_i,\bm{f}_j,\bm{f}_k), \psi_a({\bm{g}}_i,{\bm{g}}_j,{\bm{g}}_k)),
\end{split}
\end{equation}
where $\ell_3$ is the Huber loss. The angle-wise potential function $\psi_a$ is represented as
\begin{equation}\label{eq:rkd-a}
\begin{aligned}
\psi_a(\bm{f}_i,\bm{f}_j,\bm{f}_k) = <\frac{\bm{f}_i-\bm{f}_j}{\|\bm{f}_i-\bm{f}_j\|_2},\frac{\bm{f}_k-\bm{f}_j}{\|\bm{f}_k-\bm{f}_j\|_2}>
\end{aligned}
\end{equation}
where $<.,.>$ is the dot-product operator. The 3-order relational knowledge is transferring the relationship of training instances embedding by penalizing angular difference, which may be able to transfer more effective information due to its higher order, leading to more flexibility in training the student.

\myPara{Center-based relational knowledge.} Typically, when the order is more than 3, it will bring in the following two issues: 1)~the computational cost will increase, as we expect, and 2)~the potential function is difficult to define. A feasible way to address the first issue is reducing the example number in each mini-batch when training. Clearly, though, the relationship between an example and some other examples outside the mini-batch will be lost, reducing the sufficiency of knowledge transferred.
In this work, we propose the class-centered relational knowledge to address these issues. Toward this end, an extra set of examples $\bl{U}=\{\bl{u}_c\}_{c=1}^{C}$ is defined to describe the class centers in the output representation space, which is represented as
\begin{equation}
~\bl{u}_c=\frac{\sum_{i=1}^{|\chi^1|}\delta(l_i=c)\bm{f}_i}{\sum_{i=1}^{|\chi^1|}\delta(l_i=c)},
\end{equation}
where $\delta(l_i=c)$ is an indicator function which equals 1 if $l_i=c$ and 0 otherwise. $C$ is the total class number. In this way, each example is characterized by the feature instance of a specific class center, and the class center is represented by the average feature instances. Then, a support space is constructed by using these class centers, which can be used to create the C-order relational knowledge for an instance:
\begin{equation}\label{eq:rkd-4}
\begin{split}
\mc{L}_{C}=\sum_{\bm{x}_i\in{{\chi}^{1}}, \hat{\bm{x}}_i\in{{\hat{\chi}}^{1}}} \sum_{c=1}^{C} \ell_2(\psi_e(\bm{f}_i,\bl{u}_c),\psi_e({\bm{g}}_i,{\bl{u}}_c)),
\end{split}
\end{equation}
where $\psi_e$ measures the $\mc{L}_{2}$ distance between two feature instances, having $\psi_e(\bm{f}_i,\bl{u}_c)=\|\bm{f}_i-\bl{u}_c\|_2$. In this way, this high order relation can be converted into a group of 2-order relations, which can be addressed efficiently.

\myPara{Total distillation loss.}
With the hybrid order relations, the knowledge to be transferred can involve the relations in various levels, that is individual-level, pair-level, triplet-level and group-level knowledge. Finally, the total distillation loss is the weighted sum of these four losses:
\begin{equation}\label{eq:rkd-5}
\begin{split}
\mc{L}=\mc{L}_{1}+\alpha\mc{L}_{2}+\beta\mc{L}_{3}+\gamma\mc{L}_{C}.
\end{split}
\end{equation}
where, $\alpha$, $\beta$ and $\gamma$ are the tuning factors to balance the effects of different order relational knowledge.

\subsection{3.3~~Two-Stage Knowledge Transfer}
As shown in Fig.~\ref{fig:framework}, the knowledge transfer includes two stages.
In the first cross-structure distillation stage, the input of the teacher and assistant network is high-resolution images. The objective is to alleviate structure redundancy like general process of many knowledge distillation approaches \cite{2015arXiv150302531H,romero2015iclr}. After cross-structure distillation, there still exist redundancy in the resolution, meaning that a low-resolution image can still be recognizable when its corresponding high-resolution images are learned adequately. In the second cross-resolution step, we use the assistant network as a new teacher to guide the training of low-resolution student network that has the same structure as assistant network. The objective is to reduce the information loss due to resolution degradation.

\section{4~~Experiments}
To verify the effectiveness of our proposed approach, we conduct comprehensive experiments on three typical low-resolution image recognition tasks, including low-resolution image classification, low-resolution face recognition and cross-resolution metric learning.

\subsection{4.1~~Low-resolution Image Classification}
First, we check low-resolution image classification task on the challenging CIFAR100 benchmark.
The CIFAR100 dataset consists of 60K 32$\times$32 images of 100 classes. Among these images, 50K images are used for training and the remaining 10K images for testing. To benchmark our models, we further make the comparisons between four state-of-the-arts: 1)~\textbf{KD}~\cite{2015arXiv150302531H} serves as baseline that uses the final output class probabilities as soft targets to supervise the training of student, 2)~\textbf{FitNet}~\cite{romero2015iclr} uses the intermediate features as the knowledge and transfers to the student, 3)~\textbf{Attention}~\cite{attention2017iclr} also uses feature instances as the knowledge that is enhanced with attention module, and 4)~\textbf{RKD} transfers the 2-order and 3-order relational knowledge to the student whose input resolution is the same as teacher. For all the settings, we use the cross-entropy loss at the final loss in addition. For both the teacher and the student, we remove fully-connected layer(s) after the final pooling layer and append a single fully-connected layer as a classifier. The teacher is achieved by training in original 32$\times$32 training dataset with ResNet50, while the student is trained in the corresponding dataset with a reduced resolution of 16$\times$16 by using a simpler ResNet18 network. For extensive comparisons, we also conduct the experiments with the same input resolution and by using another VGG11 network~\cite{vggnet2015} for student (OUR-\textbf{S}-\textbf{N}, where \textbf{S}$\in\{16,32,...$\} stands for the resolution and \textbf{N}$\in\{$R,V$\}$ is the ResNet or VGGNet), resulting in four models: OUR-16-R, OUR-16-V, OUR-32-R and OUR-32-V. Additionally, `L' and `S' indicate the input resolution that is lower than the teacher and the same as teacher, respectively. In this case, OUR-S-R and OUR-S-V can be considered as the assistants.

\begin{table}[htbp]
  \centering
  \caption{Classification accuracy on CIFAR100 benchmark.}
    \begin{tabular}{l|c|c}
    	    \hline
    Model      & Resolution & Accuracy ($\%$) \\ \hline
    FitNet     & 32$\times$32      & 70.08 \\
    Attention  & 32$\times$32      & 72.68 \\
    RKD        & 32$\times$32      & 72.97 \\
    KD         & 32$\times$32      & 74.26 \\
    OUR-32-V   & 32$\times$32      & 73.54 \\
    OUR-16-V   & 16$\times$16      & 69.87 \\
    OUR-32-R   & 32$\times$32      & \textbf{76.68} \\
    OUR-16-R   & 16$\times$16      & \textbf{74.65} \\ \hline
    Teacher    & 32$\times$32      & 77.76 \\
    \hline
    \end{tabular}%
  \label{tab:cifar100}%
\end{table}%

The experimental results are shown in Tab.~\ref{tab:cifar100}. First, it shows that the classification accuracy consistently decreases after the data resolution is reduced from 32$\times$32 to 16$\times$16, as we expect. Specially, thank to the step-wise knowledge transfer from the teacher to the student with the help of assistant, the accuracy drop of at an acceptable level, \eg, $3.11\%$ from $77.76\%$ to $74.65\%$, while the model get smaller. Second, by using OUR-S-R with $76.68\%$ accuracy as the assistant, the student OUR-L-R achieves an accuracy of $74.65\%$, which is higher than the student OUR-L-V that uses an lower-accurate OUR-S-V as assistant. It implies that a higher-accurate assistant may lead to better student. Third, our low-resolution student OUR-L-R gives a higher accuracy than other state-of-the-art distillation approach, demonstrating the effectiveness of our approach.

\subsection{4.2 Low-resolution Face Recognition}
After the promise achieved in general image classification task, we focus on low-resolution face recognition task that is very helpful in many real-world applications, \eg, recognizing low-resolution surveillance faces in the wild. In our experiments, the teacher uses a recent state-of-the-art face recognizer VGGFace2 with ResNet50 structure. The student models use ResNet34 network and are trained in UMDFaces~\cite{Bansal2016UMDFaces} that is collected from Internet and serves as the high-resolution face dataset.  Then, the trained student models are used to evaluate face verification task on LFW benchmark and face identification task on UCCS dataset~\cite{Bansal2016UMDFaces}. LFW benchmark contains 6K pairs where 3K positive pairs have the same identities and the remaining is 3K negative pairs. UCCS contains 16,149 images in 1,732 subjects in the wild condition. We follow the setting as~\cite{ge2019tip}, randomly select a 180-subject subset, and randomly separate the images into the 3,918 training images and 907 testing images according to a ratio of about 4:1, and report the results with the standard top-K error metric. In order to verify the validity of our low-resolution models, we emphatically check the accuracy when the input resolution is 32$\times$32 and 16$\times$16. In the following, we report these results.

\myPara{Face verification on LFW.} In this experiment, we conduct the comparisons with 12 state-of-the-art face recognition models, including 6 models working at normal resolution (DeepFace~\cite{taigman2014closing}, DeepID~\cite{sun2014deep1}, DeepID2~\cite{sun2014deep2}, FaceNet~\cite{schroff2015facenet},VGGFace~\cite{parkhi2015deep} and ArcFace~\cite{deng2019arcface}), 6 models working at low resolution (MobileID~\cite{luo2016face}, SphereFace~\cite{liu2017sphereface}, ShiftFaceNet~\cite{wu2018shift}, CosFace~\cite{wang2018cosface}, VGGFace2, and SKD~\cite{ge2019tip}).
For the 6K face pairs, we extract the facial features for similarity comparison. With a pre-set threshold, each faces pair is determined to have the same identity if the similarity of the two faces is greater than the threshold and different identity otherwise. The verification accuracy is reported as the percentage of the pairs that are correctly determined. The results are listed in Tab.~\ref{tab:lfw}.

From the results, we can find that the normal face recognition models always achieve a high accuracy but cost much more parameters. For the DeepID and DeepID2 models which work at the medium resolution setting, they integrate dozens of models to achieve good results, leading to $2\times$ and $1.5\times$ parameters to our models respectively. When working at much lower resolution and costing much less parameters, our 32$\times$32 low-resolution model OUR-32-R delivers a competitive accuracy of $93.83\%$, that is only a $5.99\%$ or $5.70\%$ drop against the best ArcFace or its teacher VGGFace2, but with a $3.5\times$ or $7\times$ resolution degradation.

Moreover, we compare with one most recent state-of-the-art SKD that works at a resolution of $32\times32$ or $16\times16$. It selectively distilled knowledge from VGGFace~\cite{parkhi2015deep} as teacher and delivered the recognition accuracy of $89.72\%$ or $85.87\%$ in the resolutions of $32\times32$ or $16\times16$, respectively. By contrast, our models OUR-32-R and OUR-16-R takes VGGFace2 as teacher, achieves the accuracy of $93.83\%$ and $90.03\%$, surpassing an accuracy of $4.11\%$ and $4.16\%$, respectively. It shows the great stability and performance of our approach on low-resolution face recognition. We suspect the main reason arises from that a better teacher as supervision and richer relational knowledge may enable better feature representation.

\begin{table}[htbp]
  \centering
  \caption{Face verification results on LFW. Our student models achieve good accuracy when working at lower resolution and costing much less parameters.}
    \begin{tabular}{c|c|c|c|c}
    	\hline
    Model & Acc.(\%) & Resolution & \#Para & Year \\
    \hline
    DeepFace & 97.35 & 152$\times$152 & {120M} & 2014 \\
    DeepID & 97.45 & 39$\times$31 & {17M} & 2014 \\
    DeepID2 & 99.15 & 55$\times$47 & {10M} & 2014 \\
    FaceNet & 99.63 & 96$\times$96 & {140M} & 2015 \\
    VGGFace & 98.95 & 224$\times$224 & {138M} & 2015 \\
    MobileID & 98.37 & 55$\times$47 & {2M} & 2016 \\
    SphereFace & 99.42 & 112$\times$96 & {37M} & 2017 \\
    ShiftFaceNet & 96    & 224$\times$224 & {0.78M} & 2018 \\
    CosFace & 99.73 & 112$\times$96 & {37M} & 2018 \\
    VGGFace2 & 99.53 & 224$\times$224 & {26M} & 2018 \\
    ArcFace & 99.82 & 112$\times$112 & {37.8M} & 2019 \\
    SKD   & 89.72 & 32$\times$32 & {0.79M} & 2019 \\
    OUR-32-R   & \textbf{93.83} & 32$\times$32 & {7.8M} & - \\
    SKD   & 85.87 & 16$\times$16 &  {0.79M} & 2019 \\
    OUR-16-R   & \textbf{90.03} & 16$\times$16 & {7.8M} & - \\
    \hline
    \end{tabular}%
  \label{tab:lfw}%
\end{table}%

\begin{table}[htbp]
  \centering
  \caption{Face identification results on UCCS. Our model delivers lower error rate (\%).}
    \begin{tabular}{c|c|c|c}
    	    \hline
    Model & resolution & Top-1 & Top-5 \\
    \hline
    VLRR  & 16$\times$16 & 40.97 & 22.35 \\
    SKD   & 16$\times$16 & 32.75 & 18.3 \\
    Our-16-R   & 16$\times$16 & \textbf{22.19} & \textbf{10.23} \\
        \hline
    \end{tabular}%
  \label{tab:uccs}%
\end{table}

\myPara{Face identification on UCCS.} In the experiment, we make comparisons with two state-of-the-arts (VLRR~\cite{wang2016studying} and SKD~\cite{ge2019tip}) on the low-resolution face identification task by using the challenging UCCS dataset. We report the results with $16\times16$ resolution models.
As shown in Tab.~\ref{tab:uccs}, the VLRR model reported the best error rates of $40.97\%$@top-1 and $22.35\%$@top-5, while SKD achieves $58.65\%$@top-1 and $22.71\%$@top-5 error rates. Compared with them, our model achieves better results ($22.19\%$@top-1 and $10.23\%$@top5 error rates), showing better performance in recognizing low-resolution faces.

\begin{figure}[t]
	\centering
	\includegraphics[width=1.0\linewidth]{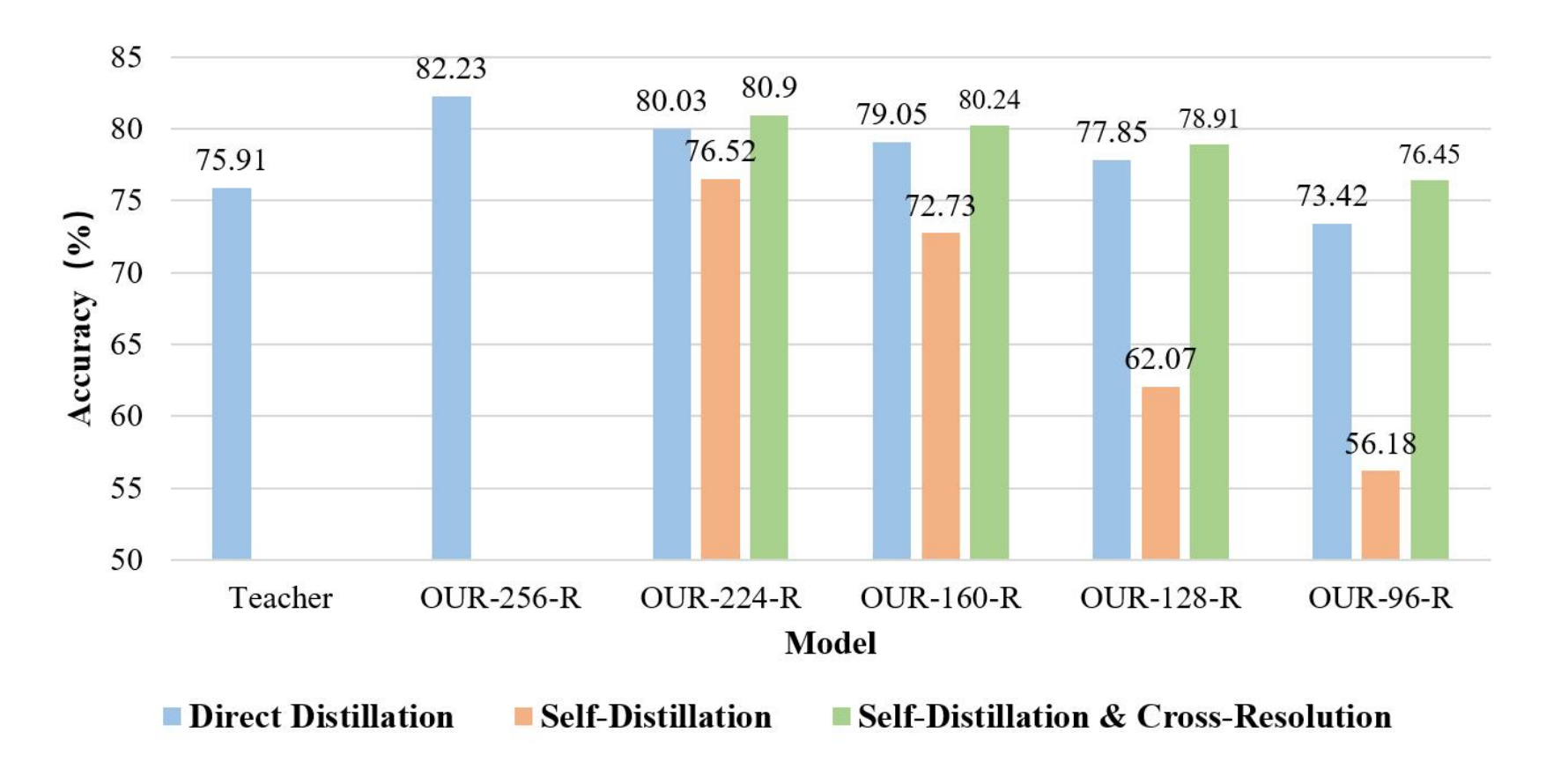}
	\caption{Results on cross-resolution metric learning.}
	\label{fig:self}
\end{figure}

\subsection{4.3~~Cross-Resolution Metric Learning}
Beyond the low-resolution image classification and face recognition tasks, we further evaluate our approach on low-resolution metric learning task which learns the similarity of objects under the resolution degradation. In the experiment, we mainly carry out cross-resolution measurement of interest correlation under three benchmarks:1)~Cars196 contains 16,185 car images in 196 classes, 2)~CUB2011 is an extended version of CUB200 that is a challenging dataset of 200 bird species, and 3)~SOP (Stanford Online Products) is a product dataset having 120,053 images of 22,634 classes.
The input resolution is normalized to 256$\times$256 for teacher and ranges from 256$\times$256 to 96$\times$96 for students.

Our main purpose is to verify the performance of our model in cross-resolution metric learning. In the experiments, We use ResNet50 as teacher and ResNet18 as student, and then conduct the experiments to show the effect of using original data to train teachers and different degraded resolution of images to feed students. The results are shown in Tab.~\ref{tab:metric}, which are compared with several state-of-the-arts: including AT~\cite{attention2017iclr}, PKD~\cite{passalis2018learning}, DarkRank~\cite{chen2018darkrank} and MKD~\cite{yu2019learning}).

The experimental results show that although the performance of student network will be limited by the resolution gap between teacher and student input, overall, our student is better than teacher. Specifically, when the resolution of the student is maintained at more than one fourth of the resolution of the teacher, that is 128$\times$128 on Cars196 dataset, our approach shows good adaptability, even the accuracy increases about 2\% from the accuracy of our teacher. On the CUB-200-2011 and SOP dataset, the performance of students is slightly worse, but still better than the best results of other approaches. These results are based on the premise that our teacher is simpler than their teacher. That is to say, we have adopted simpler student to achieve better performance in cross-resolution metrics learning.

\begin{figure}[t]
	\centering
	\includegraphics[width=1.0\linewidth]{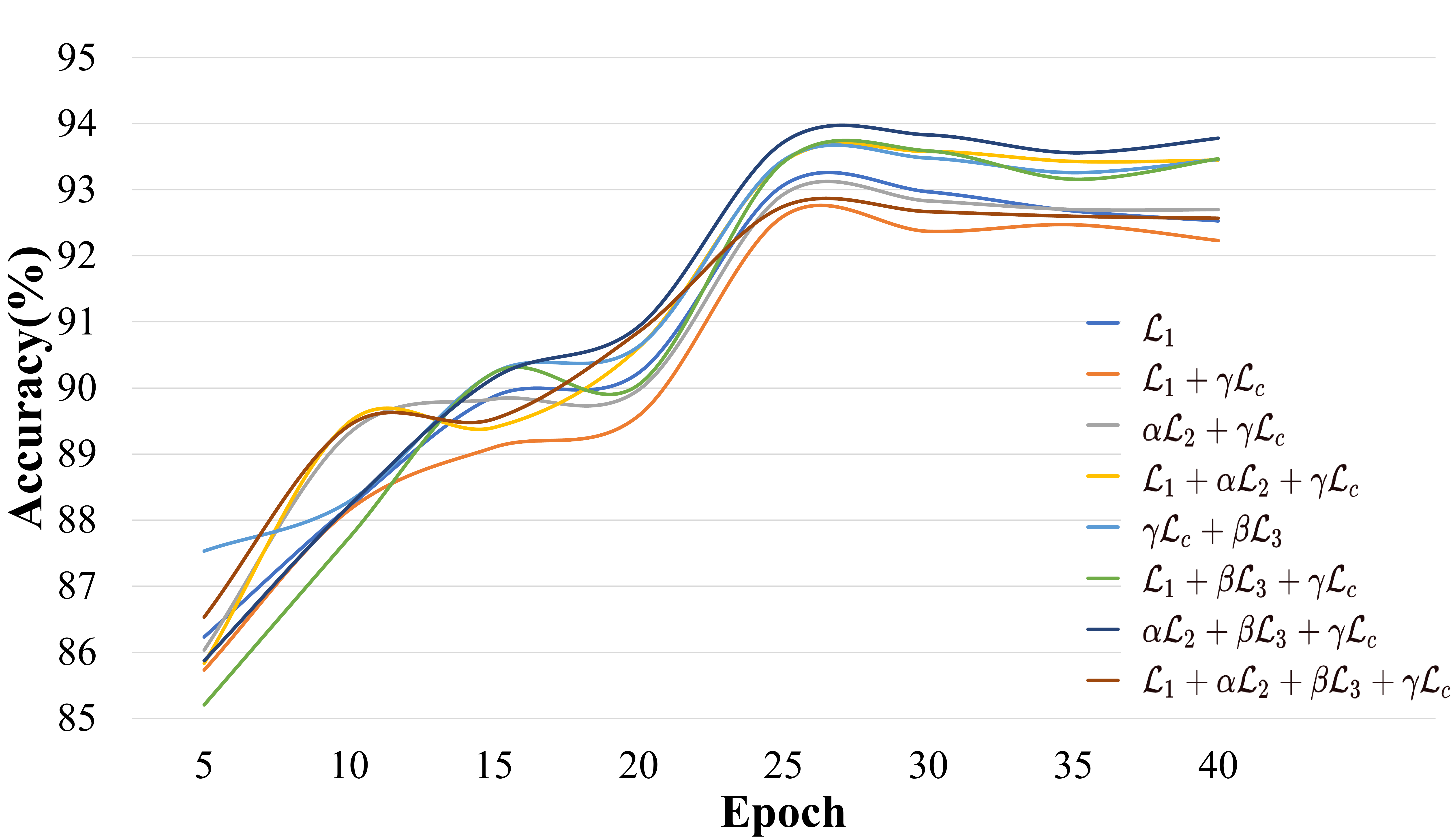}
	\caption{The effects of various order relational knowledge.
	}
	\label{fig:loss}
\end{figure}

For cross-resolution metric learning, we use ResNet50 as teacher with the input of high-resolution images, and ResNet18 as assistant. Then we use the assistant to supervise the training of student that also uses ResNet18 but takes low-resolution images as the input. From the experimental results in Fig.~\ref{fig:self}, it can be seen that the experimental results have been greatly improved by cross-resolution distillation. If we directly distill the student after cross-resolution distillation, it will decrease the accuracy of model.
In addition, it shows that the performance of student is getting slightly worse than our student as the resolution gap between the student and teacher. Specifically, when the input resolution of the student is maintained at more than one fourth of the resolution of the teacher, that is 96$\times$96, our approach shows good adaptability.

We speculate that the reason is that if the image resolution is enough to ensure the amount of information needed for recognition, self-distillation without cross-resolution through the same network structure can well retain the information needed to learn. In our method, the first non-cross-resolution distillation helps to retain more information. The main advantage of the subsequent cross-resolution self-distillation is that the network is the same, although the resolution is lost, the more information may be learned. However, direct cross-resolution distillation is not conducive to this kind of knowledge transfer.

\begin{table}[htbp]
  \centering
  \caption{Results of metric learning on three datasets.}
    \begin{tabular}{c|c|c|c}
    	    \hline
         Model & \multicolumn{1}{l}{Cars196} & \multicolumn{1}{l}{CUB2011} & \multicolumn{1}{l}{SOP} \\
              \hline
    PKT   & 46.9  & 53.1  & - \\
    DarkRank & 74.3  & 56.2  & - \\
    MKD & 76.6  & 58.0  & 68 \\
    MKD+AT & 76.4  & 58.1  & - \\
    \hline
    OUR-256-R & {82.23} & {57.82} & {68.13} \\
    OUR-224-R & {80.03} & {57.14} & {69.90} \\
    OUR-160-R & {79.04} & {57.19} & {68.87} \\
    OUR-128-R & {77.85} & {55.37} & {67.66} \\
    OUR-96-R  & {73.41 } & {53.53} & {66.42} \\
    \hline
    Student(ResNet18) & 46.70  & 51.70  & 61.70 \\
    Teacher(ResNet50) & 75.91 & 58.96 & 72.62 \\
    ResNet101         & 74.80  & 58.90  & 69.50 \\
    \hline
    \end{tabular}%
  \label{tab:metric}%
\end{table}%

\subsection{4.4~~Ablation Study}
Generally speaking, the higher-order information will yield better knowledge transfer. But does learn more information mean that the model performance is better? In order to answer this question, we conducted a special experiment on low-resolution face recognition task with cross-structure distillation to verify the effect of the order of attention on the model. The face images are taken from UMDFaces and UCCS datasets and resized to 32$\times$32. We use ResNet50 for teacher and the extremely simplified ResNet34 for student to study the performance of knowledge transfer. In our experiments, the parameters $\alpha$, $\beta$, $\gamma$ are set to 0.02,0.01,1 respectively. The results are shown in Fig.~\ref{fig:loss}.

With the increasing order of loss functions, the model performance is indeed getting better. However, the optimal results come from $\mc{L}=\alpha\mc{L}_{2}+\beta\mc{L}_{3}+\gamma\mc{L}_{C}$  rather than the highest-order $\mc{L}=\mc{L}_{1}+\alpha\mc{L}_{2}+\beta\mc{L}_{3}+\gamma\mc{L}_{C}$. In addition, when the number of loss terms is the same, the higher the order of information concerned by the loss function itself, the better the performance is. We conclude that although the increase of the order will help to improve the performance of the student while transferring knowledge, but it is not certain. Although higher order means more information, we can not guarantee that all of these information will help improve the model. The redundancy of these information may have a negative impact on the model probably. Therefore, for a specified task, finding the optimal order is very important for knowledge distillation. In our experiments, the student models perform best when $\alpha=0.02$, $\beta=0.01$ and $\gamma=1$.

Additionally, we check the merit of two-stage knowledge transfer via cross-structure distillation and cross-resolution distillation. We find that the introduction of the assistant as bridge can improve the performance for cross-resolution recognition tasks. For example, we have achieved an accuracy improvement of 3.03\% in cross-resolution metric learning task when adding cross-structure distillation.

\section{5~~Conclusion}
In this paper, we propose a hybrid order relational knowledge distillation approach for facilitating cross-resolution image recognition tasks. The approach effectively transfers rich knowledge from high-resolution teacher to low-resolution student with the help of the assistant by carrying out cross-structure distillation and cross-resolution distillation step-wisely to remove the redundancy in the network and image resolution. Extensive experimental results of three typical cross-resolution image recognition tasks show the effectiveness of the proposed approach. In the future, we will extend it for more extensive applications.

\section{Acknowledgement}
This work was partially supported by grants from the National Natural Science Foundation of China (61772513), Beijing Municipal Science and Technology Commission Project (Z191100007119002), and National Key Research and Development Plan (2016YFC0801005). Shiming Ge is also supported by the Open Projects Program of National Laboratory of Pattern Recognition, Ant Financial through the Ant Financial Science Funds for Security Research, and the Youth Innovation Promotion Association in Chinese Academy of Sciences.


\newpage
\bibliographystyle{aaai}
\bibliography{AAAI-GeS}

\end{document}